\documentclass[letterpaper, 10 pt, conference]{ieeeconf}  

\IEEEoverridecommandlockouts                              

\usepackage{framed,multirow}
\usepackage{graphicx}
\usepackage{wrapfig}
\usepackage{multirow}
\usepackage{epstopdf}
\usepackage{tabulary}
\usepackage{hyperref}
\usepackage{mathptmx} 
\usepackage{times} 
\usepackage{amsmath} 
\usepackage{amsfonts}
\usepackage{amssymb}  
\usepackage{color}
\usepackage[bottom]{footmisc} 

\newcommand{\cblue}[1] {\textcolor{blue}{#1}}
\usepackage[T1]{fontenc}

\begin{document}
\title{Interactive Open-Ended Object, Affordance and Grasp Learning \\for Robotic Manipulation
}

\author{S. Hamidreza Kasaei, Nima Shafii, Lu\'{i}s Seabra Lopes,  Ana Maria Tom\'{e}
\thanks{
All authors are with IEETA - University of Aveiro, 3810-193, Aveiro, Portugal.
S.~Hamidreza Kasaei is also with Department of Artificial Intelligence, University of Groningen, PO Box 407, 9700 AK, Groningen, the Netherlands. Email: hamidreza.kasaei@rug.nl, \{nima lsl, ana\}@ua.pt}
}

\maketitle

%

\maketitle

\begin{abstract}
Service robots are expected to autonomously and efficiently work in human-centric environments. For this type of robots, object perception and manipulation are challenging tasks due to need for accurate and real-time response. This paper presents an interactive open-ended learning approach to recognize multiple objects and their grasp affordances concurrently. This is an important contribution in the field of service robots since no matter how extensive the training data used for batch learning, a robot might always be confronted with an unknown object when operating in human-centric environments. 
The paper describes the system architecture and the learning and recognition capabilities. Grasp learning associates grasp configurations (i.e., end-effector positions and orientations) to grasp affordance categories. The grasp affordance category and the grasp configuration are taught through verbal and kinesthetic teaching, respectively. A Bayesian approach is adopted for learning and recognition of object categories and an instance-based approach is used for learning and recognition of affordance categories. An extensive set of experiments has been performed to assess the performance of the proposed approach regarding recognition accuracy, scalability and grasp success rate on challenging datasets and real-world scenarios.

\end{abstract}

\section{Introduction}

Service robots typically use a perception system to perceive the world. The perception system provides valuable information that the robot has to consider for interacting with users and environments. To assists humans in various daily tasks, a robot needs to know which kinds of objects exist in a scene, where they are and how to grasp and manipulate objects in different situations. For instance, consider a robotic task such as pouring juice from a juice-box to a mug. Such tasks consist of two phases: the first is the perception of the object (i.e., detect, localize and recognize objects) and the second is the planning and execution of the manipulation task. 

Previously, robots broadly employ static perception systems to perform object detection and manipulation tasks. The knowledge of robots is fixed, in the sense that the representation of the known object categories or grasp templates does not change after the training stage. In open-ended domains the set of categories to be learned is not predefined and it is not feasible to assume that one can pre-program all necessary object categories. Instead, robots should learn autonomously from novel experiences, supported by the feedback from human teachers. This way, it is expected that the competence of the robot increases over time.
\begin{figure}[t]
	\centering
	\includegraphics[width=\linewidth]{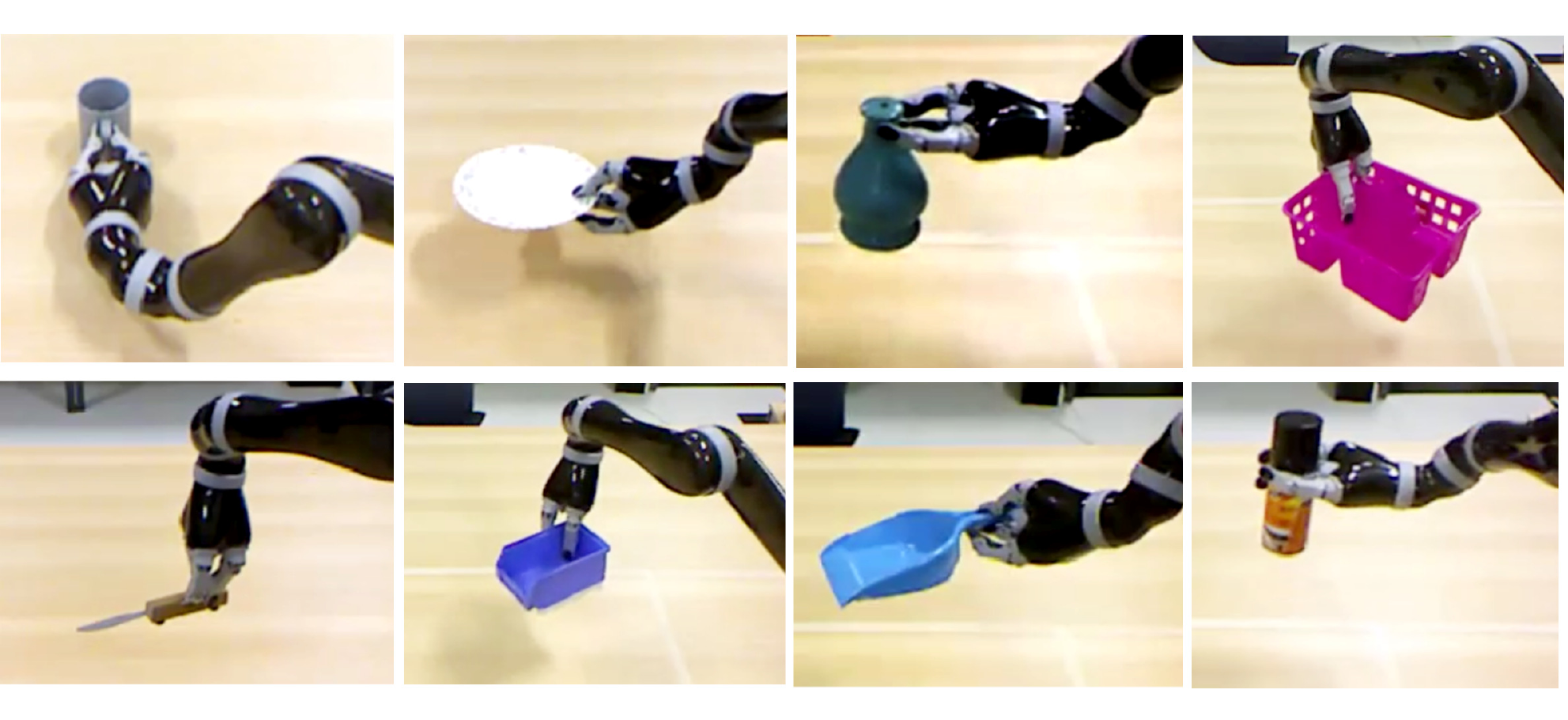}
	\vspace{-8mm}
	\caption{Eight examples of affordance detection results: given the partial point cloud of an object, we simultaneously detect the object category label, pose, and its grasp affordance.}
	\label{fig:grasp_affordance}
	\vspace{-5mm}
\end{figure}
In this paper, we approach object perception and manipulation from a long-term perspective and with emphasis on open-endedness, i.e. not assuming a pre-defined set of categories.  
As an example, consider a cutting task. If the robot does not know \emph{what} is a `\emph{Knife}', it may ask a user to show one instance and demonstrate \emph{how} to grasp a knife to execute such a task. Such situations provide opportunities to collect training instances from actual experiences of the robot and the system can incrementally update its knowledge rather than retrain from scratch when a new instance is added or a new category is introduced. In particular, we propose a complete robotic system for both open-ended object category and object affordance learning and recognition in a unified manner. We previously showed how to conceptualize tasks using experience-based robot task learning and planning \cite{mokhtari2016experience}. To the best of our knowledge, there is no other framework jointly tackling 3D object category and object affordance learning in an open-ended manner.  Fig \ref{fig:grasp_affordance} shows eight examples of our approach. We have also tried to make this framework easy to integrate into other robotic systems. 
\section{Related Work} 
\label{sec:related_work}
Over the past decade, several projcts have been conducted to develop robots to assist people in daily tasks. Jain et al.~\cite{jain2010assistive} presented an assistive mobile manipulator named EL-E that can autonomously pick objects from a flat surface and deliver them to the users. Other examples of service robot platforms that have demonstrated perception and action coupling include Rosie~\cite{beetz2011robotic}, HERB~\cite{srinivasa2010herb}, ARMAR-III~\cite{vahrenkamp2010integrated} and Walk-Man~\cite{tsagarakis2017walk}. These works are similar to ours in that they integrated perception and manipulation for pick and place operations. However, there are some differences: their vision systems are designed for detecting a set of predefined objects, while our system incrementally learns new categories through time. Furthermore, because they assumed a predefined set of objects, they computed how to grasp objects in an off-line manner or used a data-driven grasp approach~\cite{kasaei2018towards}\cite{kasaei2016object}. In our approach, grasping must handle a variety of objects never seen before.

In addition to the mentioned robotic systems, several works address the object category and the object affordance learning separately. Zero-shot~\cite{akata2014good}, low-shot~\cite{hariharan2017low}, and open-ended~\cite{kasaeiNips2016}\cite{Oliveira2016614}\cite{chauhan2015experimental}\cite{oliveira2015concurrent}\cite{kasaei2018coping} learning approaches have recently received significant attention from the machine learning and computer vision communities. In all these methods, the set of categories to be learned is not known in advance. Open-ended learning approaches, not only incrementally update the acquired knowledge, but also extend the set of categories over time. 

Grasp learning approaches can be classified according to the type of visual input. Some approaches use RGB images captured from a single viewpoint \cite{herzog2012template}\cite{detry2013learning}\cite{do2017affordancenet}\cite{nguyen2016detecting}. However, RGB data is not suitable for acquiring sufficient 3D information for grasping~\cite{philipona2003there}. Moreover, environmental changes such as light, shadows, and reflections complicate 2D detection approaches. 
Another group of approaches mainly uses RGB-D sensors, which provide only a partial view of the object. 
Herzog et al.~\cite{herzog2014learning} have developed grasp learning approaches to grasp familiar objects, in which a template matching approach is used to recognize the grasp pose. The grasp configuration is also provided to the robot via kinesthetic teaching and grasp region templates are encoded through visual features, namely height maps. In \cite{myers2015affordance}, authors proposed two approaches for learning affordances from local shape and geometry primitives. 
A third group of approaches requires knowledge about the full object geometry \cite{kokic2017affordance}\cite{amor2012generalization}. However, in real-world scenarios, it is not possible to have the complete model of all possible objects in advance. 
In previous work, we adopted an approach based on 3D partial object views. The target object view was represented
by bag-of-words and the grasp was represented by the local shape of the object around the grasp point and a global
feature of the grasp point~\cite{shafii2016learning}. There was no clear separation between the object category and
the grasp affordance. In the present paper, we modify and extend that work by separating object recognition and
grasp affordance recognition. In addition, the representation of object views is now based on a global object
descriptor and a Bayesian learning algorithm is used.

Currently, a popular approach in object recognition and affordance detection is Deep  Learning (DL). It is now clear that when we have a fixed set of object categories and a large number of examples per category, DL approaches work impressively for both object recognition and affordance detection \cite{do2017affordancenet}\cite{nguyen2016detecting}\cite{li2016fpnn}\cite{wu20153d}. 
However, there are several limitations to use DL in open-ended domains. In general, DL approaches are incremental in nature but not open-ended, since the inclusion of novel categories enforces a restructuring in the topology of the network. 
Moreover,  DL requires long training times. 
\begin{figure}[!t]
	\hspace{-6mm}
	\includegraphics[width=1.1\linewidth]{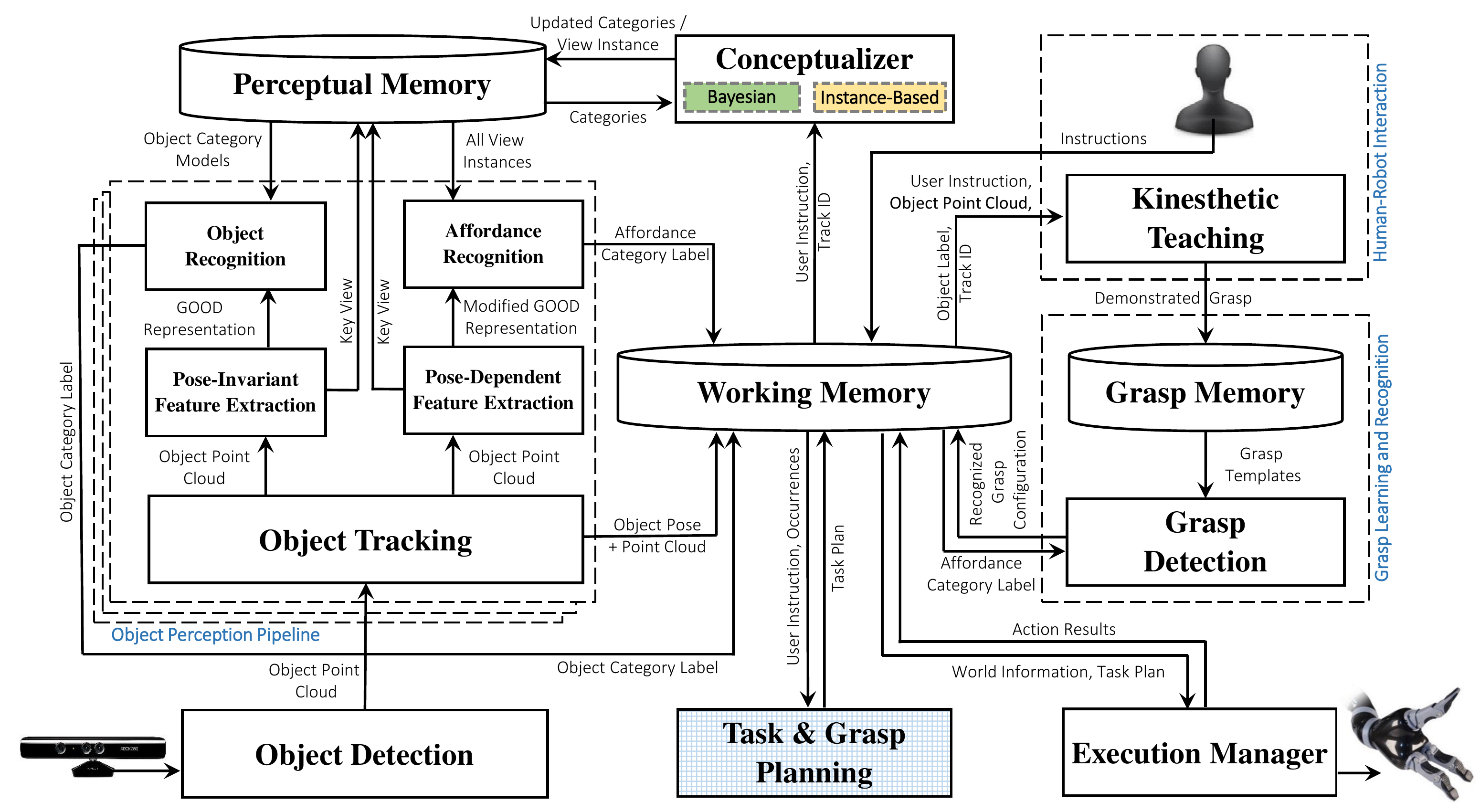}
	\vspace{-6mm}
	\caption{Overall system architecture of the proposed framework. Each box represents a module that is organized as a ROS~\cite{quigley2009ros} package and arrows signal the exchange of information between software modules.}
	\label{fig:1}
	\vspace{-5mm}
\end{figure}
\vspace{-2mm}
\section{From Object Recognition to Grasp Detection}
\label{sec:system_architecture}

The goal of this work is to concurrently learn and recognize objects as well as their associated affordances. 
Assume that there are several objects on the table and
the user asks the robot to grasp a specific object (e.g. ``\emph{grasp the mug}''). This involves several steps. First, the robot will recognize the categories of the objects on the table and locate the target object. Second, the robot will recognize the grasp affordance of the object in the current pose. Finally, given the affordance, the robot will determine a
suitable point on the object's surface for grasping and carry out the grasp action. Each of these steps is based on learned knowledge, as will be described in this section. The overall architecture of the developed system is shown in Fig.~\ref{fig:1}. In this architecture, \emph{Working Memory} is employed to support communication purposes between the different modules of the architecture.

\subsection{Human-Robot Interaction}
\label{subsec:HRI}
\begin{figure}[b]
	\vspace{-2mm}
	\centering
	\includegraphics[width=\linewidth,, trim= 0cm 0.5cm 0cm 0cm,clip=true]{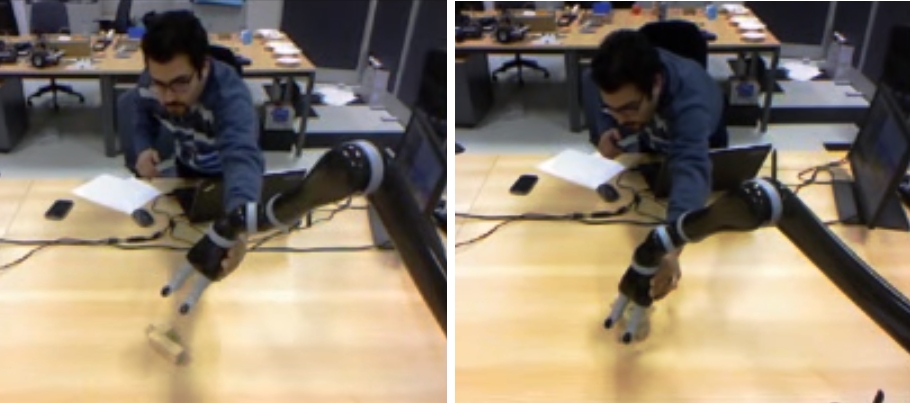}
	\vspace{-3mm}
	\caption{Kinesthetic teaching: (\emph{left}) the teacher interacts with the robot by moving the robot's gripper to a proper position; (\emph{right}) Then, the teacher demonstrates a feasible grasp for the Pentomino object to the robot.}
	\label{fig:kinesthetic_teaching}
		\vspace{-5mm}
\end{figure}
The Human-Robot Interaction (HRI) interface supports a set of actions that a teacher can use for interacting with the robot. 
In particular, the user can select an object to be target of the next action, teach the category of the target object as well as its affordance category, ask for category predictions, correct predicted categories, teach grasp configurations and ask the robot to grasp an object. Verbal interaction is used for teaching and testing categories and kinesthetic teaching is used for teaching grasps.
As shown in Fig.~\ref{fig:kinesthetic_teaching}, an instructor teaches an appropriate end-effector position and orientation using the robot's compliant mode\footnote{An example video is online at: \href{https://youtu.be/HoEjJJOynmY}{\cblue{https://youtu.be/HoEjJJOynmY}}}. 
When the agent fails to recognize the category of an object correctly, the teacher can give a correction. Therefore, at the most basic level of interaction, the interface allows the user to perform the following actions:

\begin{itemize}
\item \textbf{Select:} point to the target object or select its \emph{TrackID} from a menu.

\item \textbf{Teach-category:} teach the object category or the affordance category of the selected object (each stable pose of an object on the table may map to a different affordance category).

\item \textbf{Ask-category:} inquire the category or the affordance of the target object, which the agent will predict based on previously learned knowledge.

\item \textbf{Correct-category:} if the agent could not recognize a given object or its affordance correctly, the user can teach the correct one. 

\item \textbf{Teach-grasp:} using kenesthetic teaching, teach a grasp configuration of the robotic arm to grasp the
target object.

\item \textbf{Grasp:} command the robot to grasp the target object.
\end{itemize}

The robot reacts to the actions of the user by either running the relevant learning functionalities (i.e., in the cases of teach and correct actions) or using the learned knowledge to performe the task (i.e., recognition and/or grasping).  

\subsection{Perceptual Learning and Recognition}
\label{sec:perceptual_learning_recognition}

As it is shown on the left side of Fig.~\ref{fig:1}, we first employ an object detector. Then, the object and affordance categories are predicted using the previously acquired category knowledge.
\subsubsection{Object Detection and Tracking}

We use a recently proposed method \cite{kasaei2018perceiving} in the Object Detection module. This method demonstrates good results on both isolated objects 
as well as objects in piles. A region of the given point cloud is considered as an object candidate whenever points inside the region are continuous in both the orientation of surface normals and the depth values. A region growing segmentation algorithm~\cite{henry2013patch} is also applied on medium-size hypotheses. The purpose of this algorithm is to merge the points that are close enough concerning the smoothness and color constraints. Each cluster of points will be treated as an object candidate. \emph{Object Detection} launches a new \emph{object perception pipeline} for each detected object and pushes the object's point cloud to the pipeline~\cite{kasaei2018perceiving}. \emph{Object Tracking} receives the point cloud of the detected object, computes an oriented bounding box and estimates the current pose of the object based on a particle filter, which uses shape and color data~\cite{Oliveira2016614} (see Fig.~\ref{fig:whashington_dataset} \emph{left}). 
As depicted in Fig.~\ref{fig:1}, the object perception pipeline has two paths, one (on the left) for object category
recognition and the other (on the right) for affordance recognition.

\subsubsection{Object Category Learning and Recognition}

Given an input object point cloud, the \emph{Pose-Invariant Feature Extraction} module computes the Global
Orthographic Object Descriptor (GOOD)~\cite{kasaei2016good} to represent the object view. GOOD is formed by concatenating the three orthographic projections of the object view in a unique and repeatable local reference frame~\cite{kasaei2016good}.
For category learning, an open-ended formulation of the Naive Bayes approach is adopted~\cite{kasaei2018towards}. Therefore, assuming each object is described by a vector $\textbf{x} = [x_1,\dots, x_n]$, each object category, $\textbf{C}_k$, is represented by a tuple:
\begin{equation}
\textbf{C}_k = \langle~N_k,~\textbf{a}_k,~\operatorname{P}(\textbf{C}_k),~[\operatorname{P}(x_1|\textbf{C}_k),\dots, \operatorname{P}(x_n|\textbf{C}_k)]~\rangle,
\end{equation}
\noindent where $N_k$ is the number of seen instances in category $k$, $\textbf{a}_k$ is a vector of bin accumulators for category $k$, $a_{ki}, i=1, \dots, n$, is the accumulation of the $i^{th}$ bin over all instances of category $\textbf{C}_k$. $\operatorname{P}(\textbf{C}_k)$ is the prior probability of category {$\textbf{C}_k$} and $\operatorname{P}(x_i|\textbf{C}_k)$ is the probability of a point falling into bin $i$ in category $k$.

The \emph{teach} and \emph{correct} actions of the user lead the robot to create a new category or to modify an existing category. In particular, whenever the user explicitly \emph{teaches} a new category,  the category is initialized using a set of views of the target object (i.e., \emph{Conceptualizer}). For simplicity, the process is formalized below assuming that each teaching action provides a single object view. The new instance, represented as a histogram $\textbf{x}^{\prime}=[x^{\prime}_1,\dots, x^{\prime}_n]$, is added to the taught category $\textbf{C}_k$.
Category initialization involves updating the total number of instances of all known categories, $N$, and initializing category specific parameters, namely the number of instances of the category, $N_k$, and the bin accumulators, $a_{ki}$:
\begin{equation}
		N \leftarrow N+1, \quad  N_k \leftarrow 1, \quad a_{ki} \leftarrow x^{\prime}_{i} \quad for\quad i=1,~\dots,~n,
\end{equation}

If the user provides corrective feedback for a known category, $\textbf{C}_k$, the category model is updated using that particular instance: 
\begin{equation}
	N \leftarrow N+1, \quad 	N_k \leftarrow N_k+1, \quad a_{ki}  \leftarrow  a_{ki} + x^{\prime}_{i},
\end{equation}

Upon each teaching action, the probabilities are updated, namely the probability of all existing categories:
\vspace{-1mm}
\begin{equation}
\operatorname{P}(\textbf{C}_k) =\frac{N_k}{N}, \quad k \in \{1, \dots m\}
\end{equation}
\noindent
where $m$ is the number of known categories up to now and the probabilities of each bin, ${x}_i$, in the category $k$, $\operatorname{P}({x}_i|\textbf{C}_k)$, and 
 $\operatorname{P}({x}_i|\textbf{C}_k)$ is updated as follows:

\begin{equation}
\label{update_p}
\operatorname{P}({x}_i|\textbf{C}_k) =\frac{a_{ki}~+1}{\sum\limits_{j=1}^{n}(a_{kj}+1)}.
\end{equation}

\noindent 
Note, the probabilities are estimated with Laplace smoothing, by adding one to each accumulator, i.e., $\operatorname{P}({x}_i|\textbf{C}_k) \neq 0$. 

To classify a given object \textbf{O}, we use Bayes rule to compute the posterior probability of each object category
and, based on that, select the category that maximizes that probability:

\begin{equation}\label{eq:muli_proj_5}
\operatorname{Category}(\mathbf{O}) = \underset{{c}_k\in~\textbf{c}}{\operatorname{argmax}}~
\operatorname{P}(\textbf{C}_k|\textbf{O}).
\end{equation}

\subsubsection{Grasp Affordance Learning and Recognition}
Grasp affordances are only loosely related to object categories. Different objects in different poses may afford the same grasp. The other way around, the same object in different poses will probably afford different grasps. Therefore, in this work affordance teaching is kept orthogonal to object category teaching. The teacher teaches the same affordance category for similar object view shapes in similar poses.

\begin{figure}[!b]
\centering
	\vspace{-4mm}
	\begin{tabular}{cc}
	\hspace{-5mm}
			\includegraphics[width=0.43\linewidth]{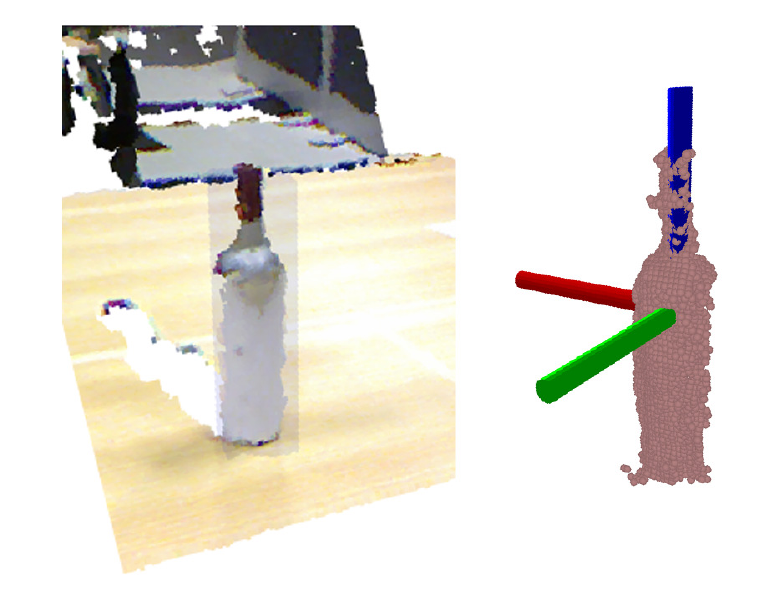}&~~
			\includegraphics[width=0.47\linewidth]{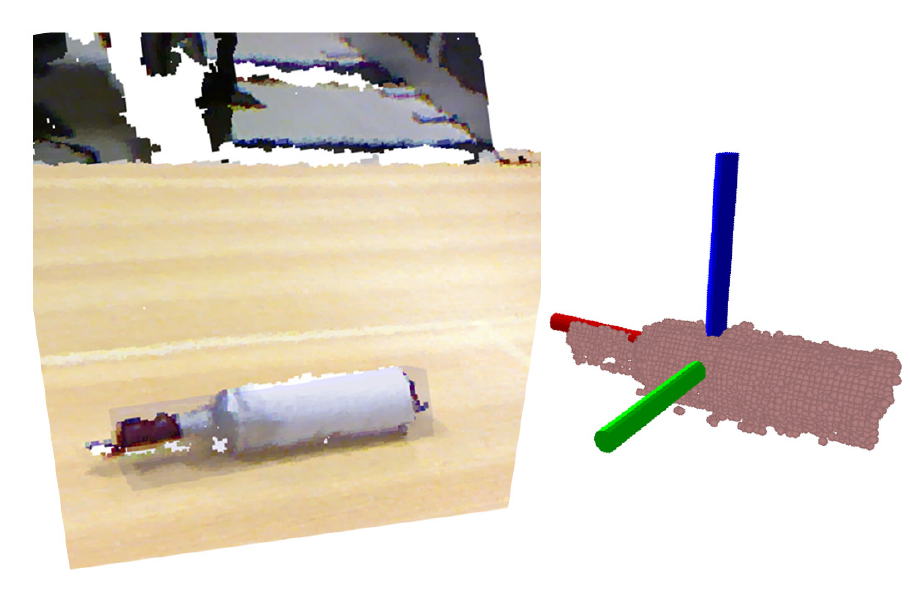} 
	\end{tabular}
	\caption{Constructing Local Reference Frames (LRF) for the bottle object in two different situations. The red, green and blue lines represent the unambiguous \textbf{X}, \textbf{Y}, \textbf{Z} axes respectively.}
	\label{fig:8}	
	\vspace{-4mm}
\end{figure}


Since grasp affordances depend on the pose of the target object, a modified version of the GOOD descriptor
\cite{kasaei2016good} is used here. 
We assume the given object is laying on a surface, e.g., a table, and therefore assign the \textbf{Z} axis to the direction that is perpendicular to the table (gravity direction). The \textbf{X} and \textbf{Y} axis must be calculated to construct the reference frame. Towards this end, we project all points of the object on the table and compute the axes of minimum and maximum variance in the horizontal plane using Principal Component Analysis (PCA). Then, the axis with maximum variance is assigned to the \textbf{X} axis. A sign disambiguation procedure is applied on the \textbf{X} as proposed for GOOD \cite{kasaei2016good}. The \textbf{Y} axis is calculated by the outer product of \textbf{Z} and \textbf{X}.
\begin{figure}[!t]
\vspace{-0mm}
\hspace{-3mm}
\begin{tabular}{cc}
		\vspace{-3mm}\multirow{2}{*}{\includegraphics[width=0.29\columnwidth, trim=-0cm 0cm 0cm 0cm,clip=true ]{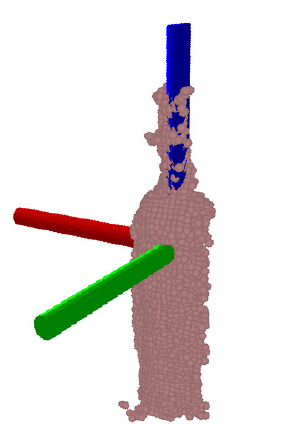}} \\ \vspace{-5mm}&\hspace{-5mm}
		\begin{tabular}{ccc}
			\hspace{-5mm}\includegraphics[width=0.2\columnwidth]{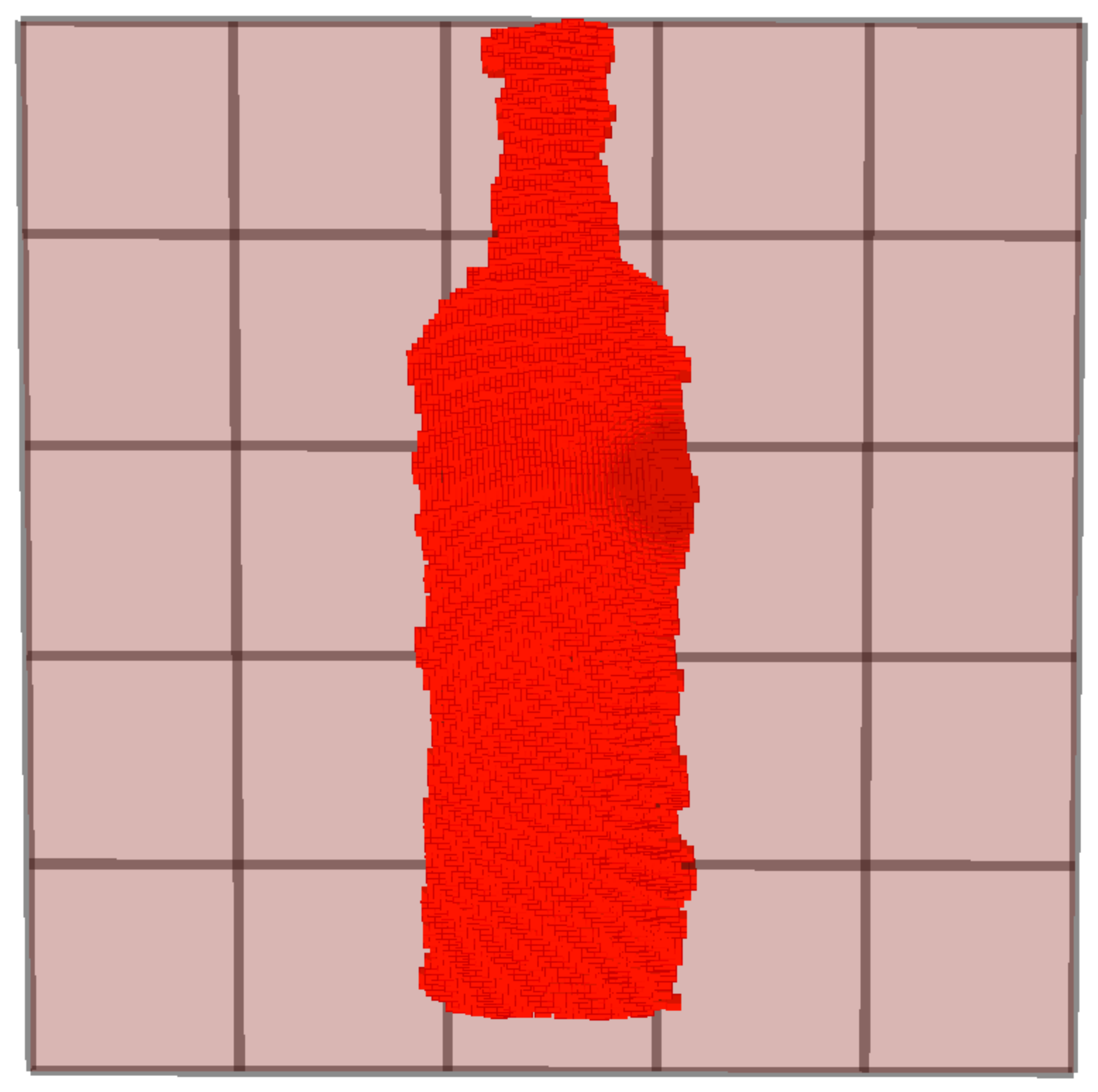}&
			\hspace{-5mm}\includegraphics[width=0.2\columnwidth]{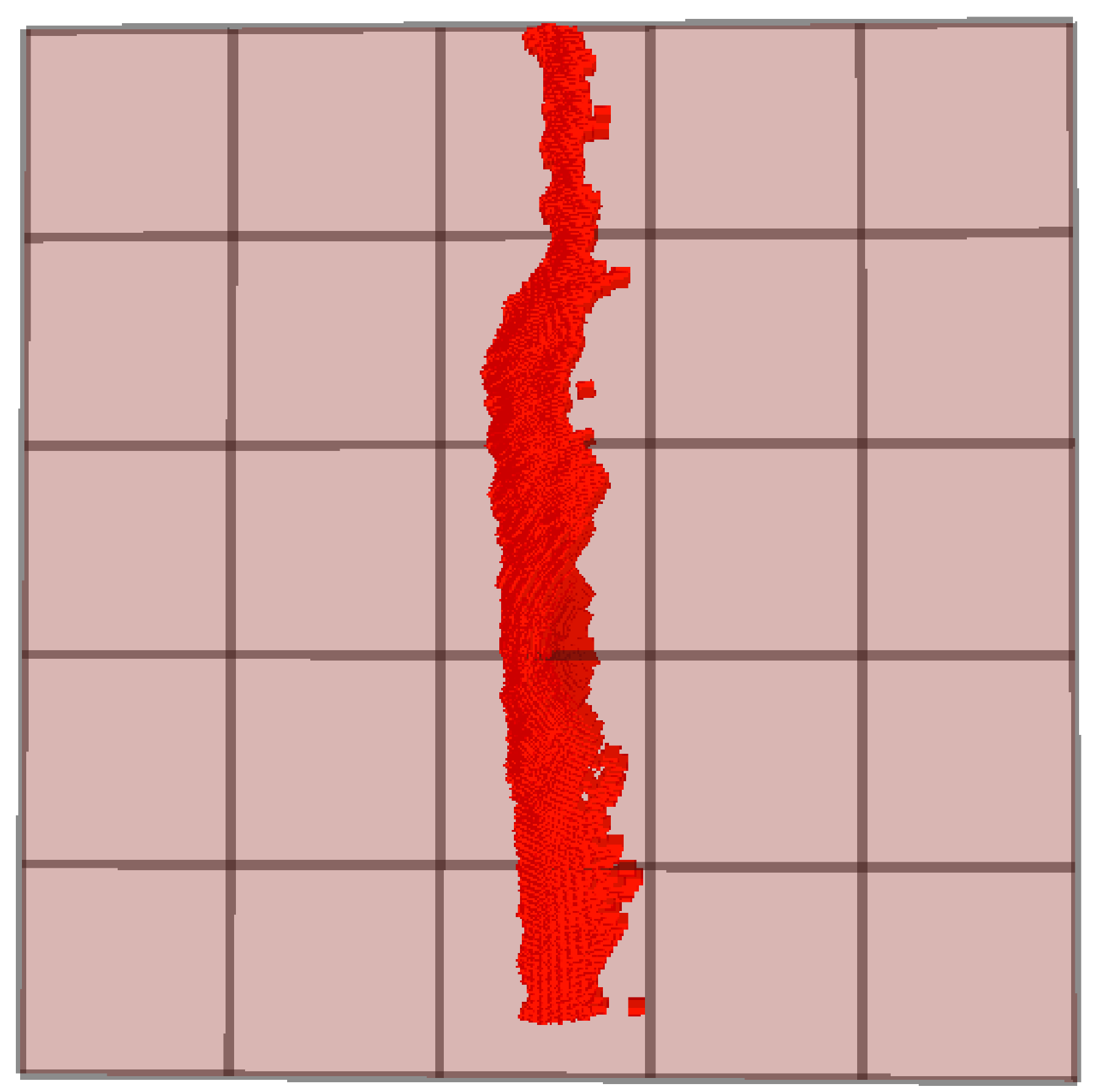}&
			\hspace{-5mm}\includegraphics[width=0.2\columnwidth]{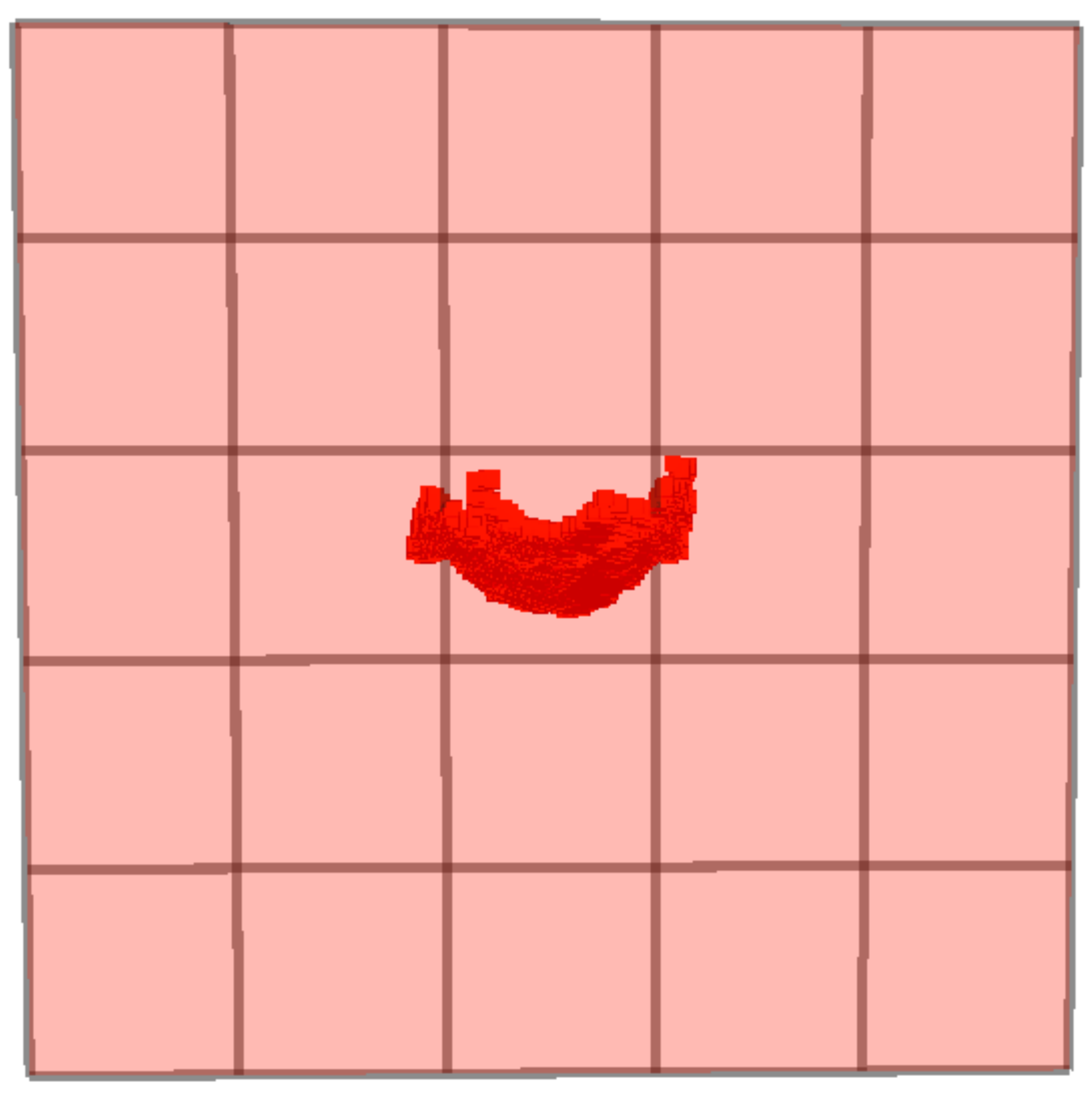} \\
			\hspace{-5mm}(\emph{b}) & \hspace{-5mm}(\emph{c}) & \hspace{-5mm}(\emph{d})\\
			\hspace{-5mm}\includegraphics[width=0.25\columnwidth]{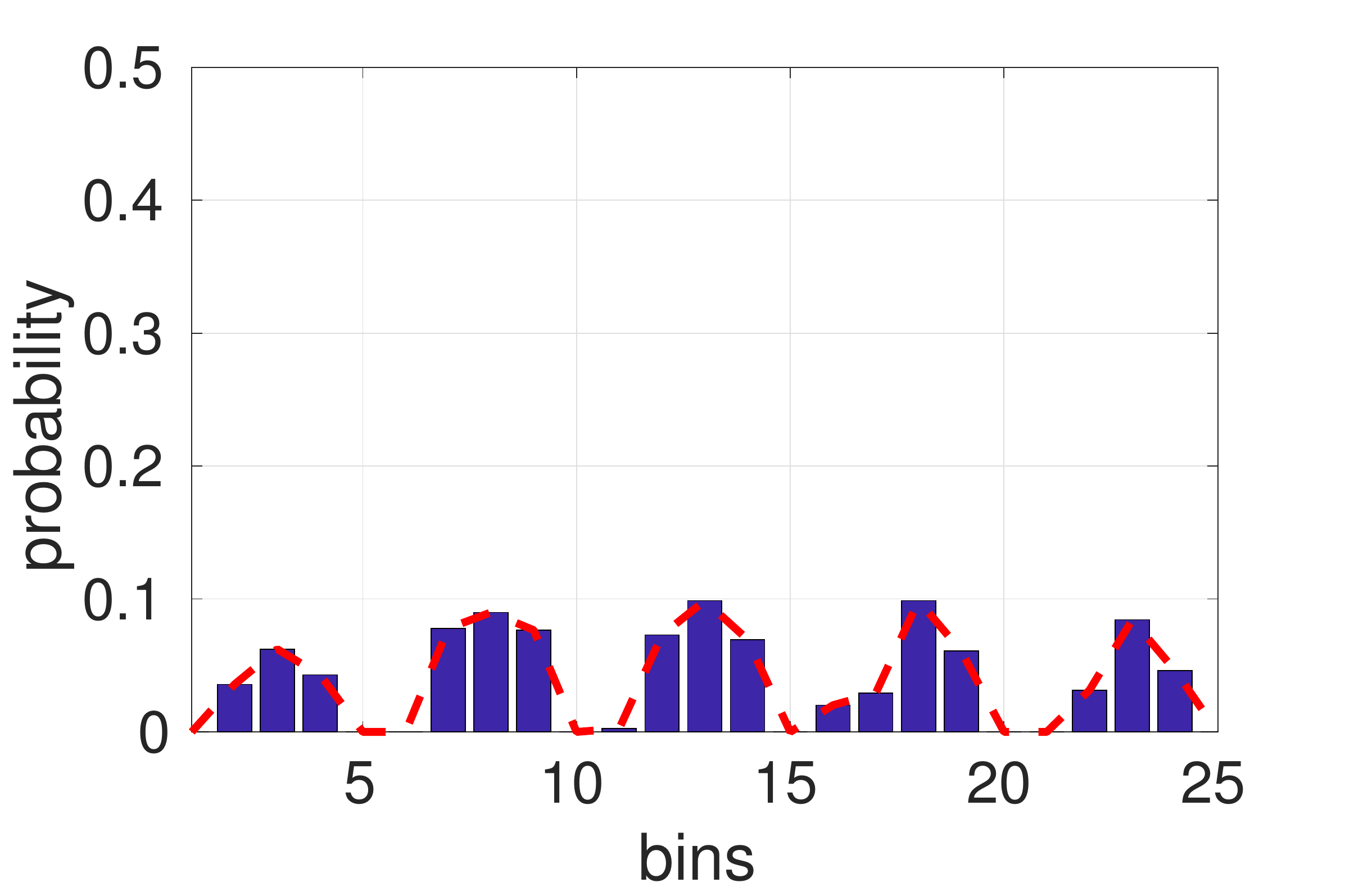}&
			\hspace{-5mm}\includegraphics[width=0.25\columnwidth]{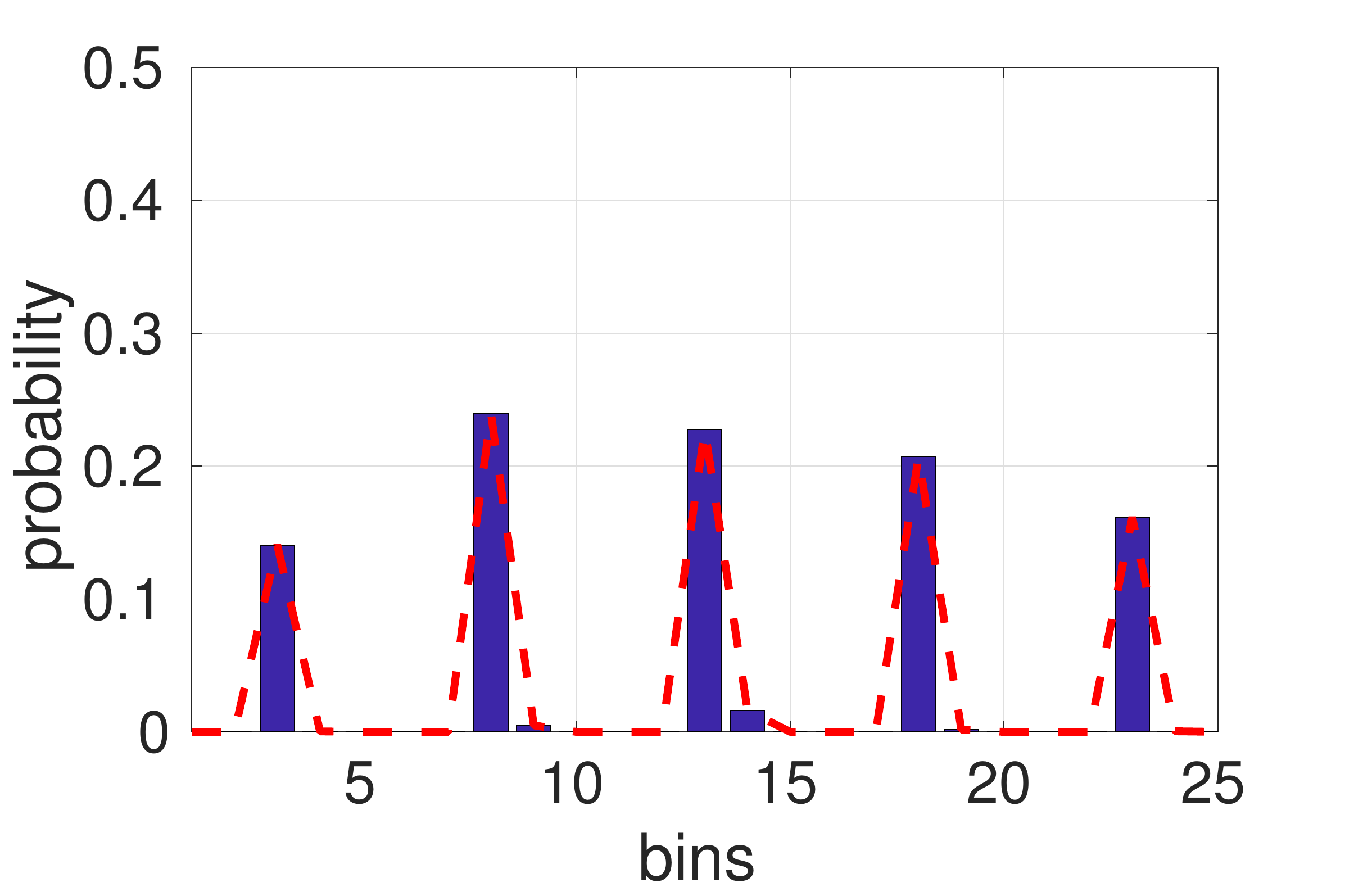}&
			\hspace{-5mm}\includegraphics[width=0.25\columnwidth]{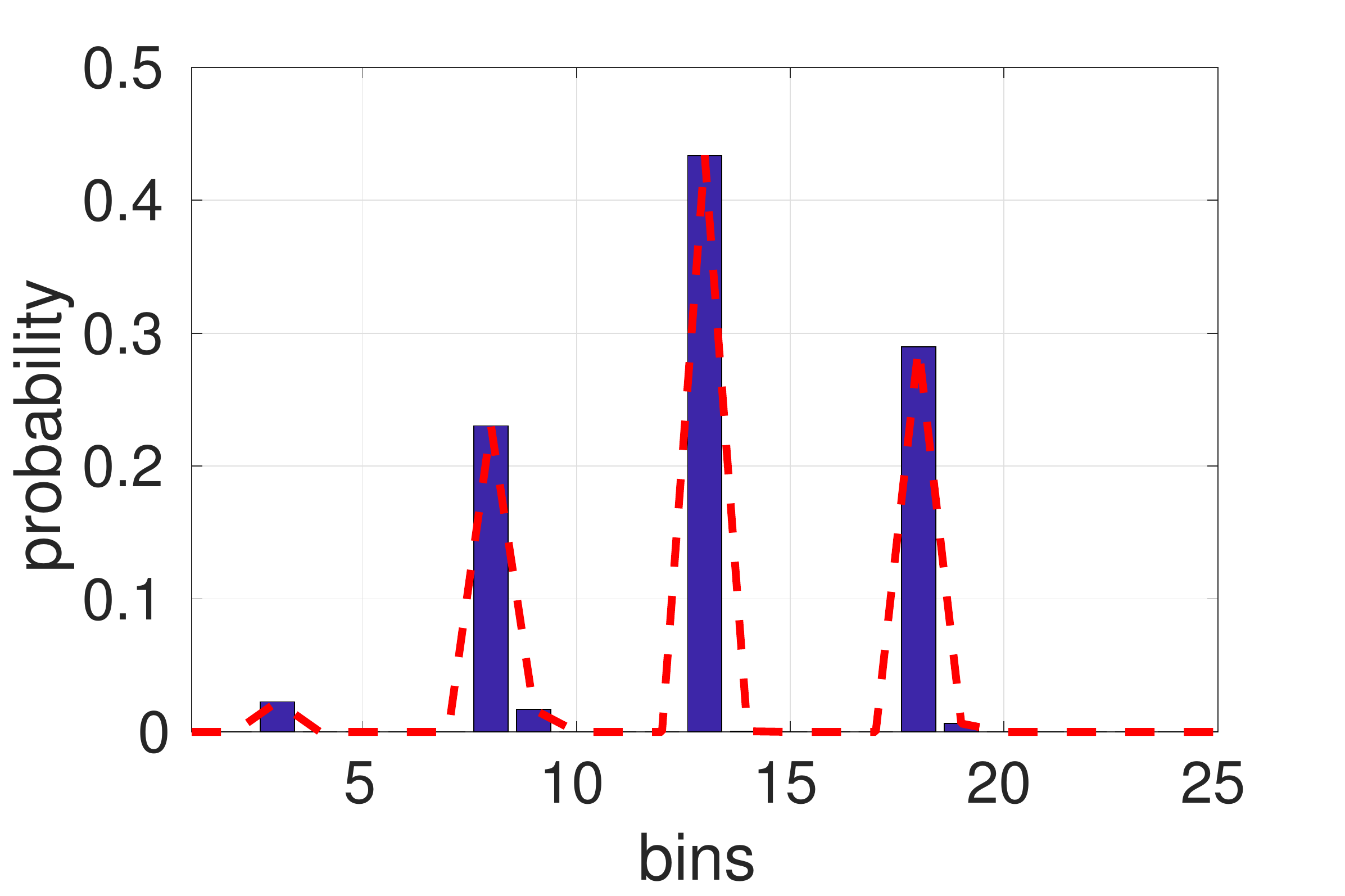} \\
			\hspace{-5mm}(\emph{e}) & \hspace{-5mm}(\emph{f}) & \hspace{-5mm}(\emph{g}) 
		\end {tabular}\\{(a)}\\
\end{tabular}
\vspace{-3mm}
\caption{An illustrative example of producing the modified GOOD shape description for a bottle object, using five bins: \emph{(a)} The bottle object and its reference frame; The red, green and blue lines represent the unambiguous \textbf{X}, \textbf{Y}, \textbf{Z} axes respectively. \emph{(b)} $XoZ$, \emph{(c)} $YoZ$ and \emph{(d)} $XoY$ projections are created. Each projection is partitioned into bins, the number of points falling into each bin is counted and three distribution matrices are obtained for the projected views; afterwards, each distribution matrix is converted to a distribution vector, (i.e. \emph{(e)}, \emph{(f)} and \emph{(g)}); The distribution vectors are concatenated, $[XoZ$, $YoZ$, $XoY]$, to form a single description.}
	\label{fig:complete_example}       
	\vspace{-5mm}
\end{figure}
As it is shown in Fig.~\ref{fig:8}, when the bottle topples on the table, an entirely different LRF is constructed compared to when the same bottle is standing on the table. In the case of the toppled bottle, the robot should grasp the bottle from the top. In the upright case, the bottle should be grasped from the side. Figure \ref{fig:complete_example} illustrates an example of the modified GOOD computation procedure for an upright bottle.

We use an instance-based learning and recognition approach~\cite{daelemans2005memory}. The advantages of instance-based approaches are that they can recognize affordances using a small number of instances and the training phase is very fast. Moreover, instance-based approaches tend to handle well heterogeneous categories. This is an important feature since objects of different categories may fall inside the same affordance category. For predicting the affordance category of the target object, the \emph{Affordance  Recognition} module first retrieves the representation of all stores instances from the \emph{Perceptual Memory} and calculates the Euclidean distance between the target object view and each of the retrieved instances. Finally, the target object is classified using the nearest neighbour rule. In our current implementation, if, for all affordance categories, the minimum dissimilarity is larger than a given threshold, the object is classified as \emph{Unknown}.

\subsection{Grasp Learning and Detection}
\label{sec:grasp_detection}

One of the main challenges is to decide which visual cues should be used as features of the taught grasp region. 
Following previous work \cite{shafii2016learning}, a combination of a local shape feature (a spin-image \cite{Johnson1999}) and a simple global feature is used. Towards this end, a key-point in the grasp region is selected based on the grasp line, i.e., a line defined by the orientation of the end-effector and passing in its center. The selected key-point is the point in the point cloud of the object that is nearest to the grasp line and also located on the surface of the object facing to the robotic arm. The spin-image \cite{Johnson1999} is computed for the selected key-point by considering the grasp region points\footnote{The parameters of the spin-image are set to: Image Width~=~8~bins, Support Length~=~0.09~m, and the surface normal area is set to 0.03~m.}. In addition, the distance of the key-point to the center of the bounding box of the object view (i.e. radius) is also computed.
Finally, the demonstrated grasp template, including the affordance category, the spin-image, the radius feature and the taught end-effector position and orientation, is stored in the \emph{Grasp Memory}.

For detecting the grasp point in the target object, the affordance category is recognized and, all the taught grasp templates with same affordance category are retrieved from the \emph{Grasp Memory}. Then, since the dimensions of the radius and the spin-image features are heterogeneous, the similarity is evaluated based on Mahalanobis distance. The grasp point is selected as the most similar template and reachable for the robot arm \cite{shafii2016learning}. 

\section{Results and Discussion}
\label{sec:results_and_discussion}
Three types of experiments were carried out to evaluate the proposed approach. 
\subsection {Open-Ended Object Category Learning and Recognition }

An evaluation protocol for open-ended learning systems was proposed in \cite{chauhan2011}\cite{Seabra2007}. The idea is to emulate the interactions of a robot with the surrounding environment over significant periods of time. We developed a \emph{simulated teacher} to follow the teaching protocol and autonomously interact with the system. The \emph{simulated teacher} repeatedly picks unseen object views of the currently known categories from a dataset, presents them to the system and estimates the recognition accuracy of the system. When accuracy exceeds a given threshold ($\tau$ = 0.67, meaning accuracy is at least twice the error rate), the teacher introduces an additional object category. This way, the system is trained online, and at the same time, the accuracy of the system is continuously estimated. In case the agent can not reach the classification threshold after a certain number of iterations (i.e., 100 iterations), the teacher infers that the agent is not able to learn more categories and terminates the experiment (breakpoint). It is possible that the agent learns all existing categories before reaching the breaking point. In such a case, it is not possible to continue the protocol, and the experiment is halted. In the reported results, this is shown by the stopping condition, ``\emph{lack of data}''. 
For the comparison, we used three other object representations approaches. Since the order of introducing categories may have an effect on the performance of the system, ten experiments were carried out for each approach.

\subsubsection{Dataset and Evaluation Metrics}
In this work, the \emph{simulated teacher} was connected to the Washington RGB-D Object Dataset consisting of 250,000 views of 300 everyday household objects, organized into 51 categories \cite{Lai2011}. We have excluded the `Ball' and `Binder' categories because of high shape similarity to the `Apple' and `Notebook' categories, respectively. Since we are using depth information, and no color or texture information, it is impossible to distinguish these categories. We have evaluated our experimental results using the main metrics introduced in previous work~\cite{chauhan2015experimental}\cite{oliveira2015concurrent}, including: (\emph{i}) the number of learned categories at the end of an experiment (TLC), an indicator of {\emph{how much the system is capable of learning}}; (\emph{ii}) the number of question/correction iterations (QCI) required to learn those categories and the average number of stored instances per category (AIC), indicators of \emph{{time and memory resources required for learning}}; (\emph{iii}) Global Classification Accuracy (GCA), an accuracy computed using all predictions in a complete experiment, and the Average Protocol Accuracy (APA), indicators of \emph{{how well the system~learns}}.

\begin{table}[!b]
\begin{center}
\vspace{-3mm}
\caption {Summary of open-ended evaluations.}
\resizebox{\columnwidth}{!}{
\begin{tabular}{ |c|c|c|c|c|c| }
\hline
Approaches & \#$\operatorname{QCI}$ & $\operatorname{ALC}$ & $\operatorname{AIC}$ & $\operatorname{GCA}$ & $\operatorname{APA}$ \\
\hline \hline
BoW\cblue{$^{\textbf{(\#)}}$}\cite{kasaei2018towards} & 1811.60 & 47.40 &14.78 & 0.69 & 0.75\\
\hline 
LDA~\cite{blei2003latent} & 900.20 &31.00 &12.25 &0.68 &0.76\\
\hline
Local-LDA\cblue{$^{\textbf{(*)}}$}\cite{kasaeiNips2016} & 1359.50 &{\textbf{49.00}} &10.01 &0.75 &0.78\\
\hline
\textbf{Our Work}\cblue{$^{\textbf{(*)}}$} & \textbf{1249.10}	& {\textbf{49.00}} &	\textbf{8.46} & \textbf{0.79}	& \textbf{0.83}\\
\hline 
\end{tabular}}
\label{table_open_ended_evaluations}
\\\scriptsize{\cblue{$^{\textbf{(*)}}$} Stopping condition was ``\emph{lack of data}''. \cblue{$^{\textbf{(\#)}}$} Stopping condition was ``\emph{lack of data}'' in 6 out of 10 experiments. }
\vspace{0mm}
\end{center}
\end{table}

\subsubsection{Results}
Table~\ref{table_open_ended_evaluations} summarizes the obtained results. One important observation is that the agent learned all 49 categories using GOOD and Local-LDA~\cite{kasaeiNips2016} and all experiments concluded prematurely due to the ``\emph{Lack of data}'' condition (indicating the potential for learning many more categories). The agent with BoW~\cite{kasaei2018towards} obtained acceptable scalability (i.e., the agent on average learned 47.50 categories). The scalability of LDA~\cite{blei2003latent} was very low (i.e., on average learned 31 categories) and its performance drops aggressively when the number of categories increases. It is also clear that the agent with GOOD stored fewer instances per category (AIC) than the other approaches. It can also be concluded that GOOD learned all categories faster than the Local-LDA approach. The agent with BoW and LDA achieved the third and fourth places respectively. By comparing all approaches, it is visible that the agent with GOOD achieved the best accuracy (i.e., 79\%) with stable performance and outperformed the other approaches by a large margin (i.e., around 4\% or more). The agent with Local-LDA also showed a promising performance and provided a good balance among all parameters. The average protocol accuracy of the agent with GOOD is also considerably higher than the other approaches (i.e., more than 5\%). It should be noted that these results should be seen in the light of the number of categories learned. For example, the BoW and the LDA approaches have average protocol accuracy (APA). However, LDA on average reached the breakpoint after the introduction of the $31^{th}$ category whereas BoW learned around 47 categories on average.
\vspace{-1mm}
\subsection {Affordance Recognition and Grasp  Detection}
\begin{figure}[!b]
	\vspace{-0mm}
	\centering
	\includegraphics[width=0.98\linewidth]{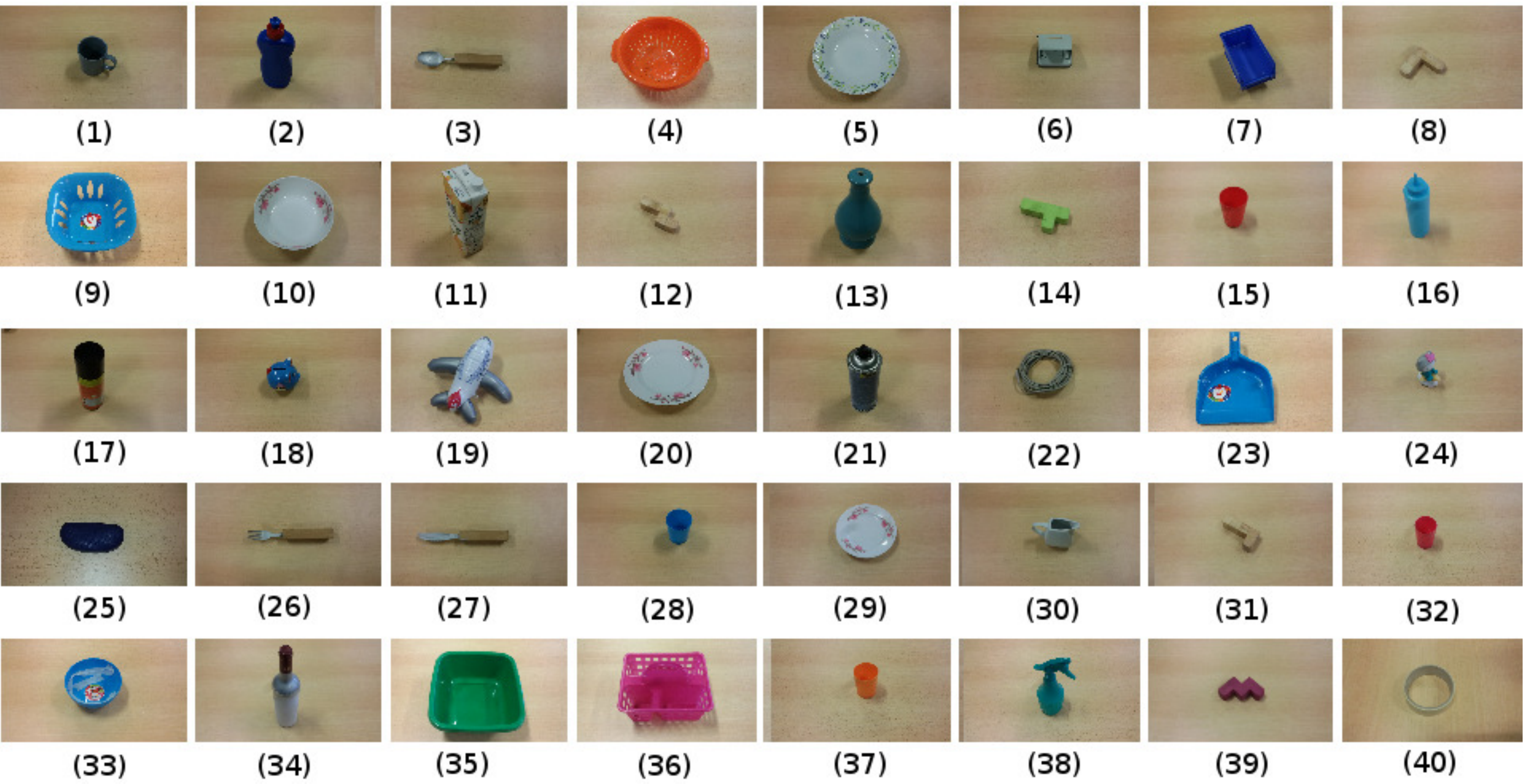}
	\vspace{-1mm}
	\caption{A set of forty household objects used to evaluate the proposed object affordance detection approach on the JACO robot.}
	\label{fig:grasp_objects}
\end{figure}
We empirically evaluate our grasping methodology using a Kinova Jaco robot. We designed a scenario in which the robot first picks-up an object and carries the object to a placing position to see if the object slips due to bad grasp or not. A particular grasp was considered a success if the robot is able to complete the pick-and-place task. In this experiment, 40 different household objects were used (see Fig. \ref{fig:grasp_objects}). We first taught how to grasp the first six objects (i.e., number 1 to 6). For each object, we taught the affordance label and the end-effector pose. For convenience, the affordance labels were the numeric identifiers of the objects.  Then, the robot tried to grasp each of the 40 objects \emph{\textbf{four times}} and the success rate was calculated. In a second round, we taught how to grasp two additional objects, namely no.~7 and no.~8, and computed the success rate to see the improvement.

\subsubsection{Grasping without affordance recognition}

In this experiment, the robot could grasp $14$ out of $40$ objects successfully in all trials (i.e.  objects 1 to 6, and 7, 10, 16, 21, 22, 24, 23, 33). The robot always failed to grasp six objects namely numbers 19, 20, 29, 28, 32 and 37, because the robot used the taught grasp template of object no.~2 (dishwashing liquid) for grasping cup-like objects instead of the grasp template taught for the object no. 1 (cup). The robot used the taught grasp template of object no.~4 (colander) for grasping some of the plate-like objects instead of using the taught template of object no.~5 (plate). The remaining 20 objects were successfully grasped in some trials, but not all. The success rate of grasping for all trials was about 58\% (93 successful trials out of 160). In the second round, by using the additional grasp templates, the robot could grasp 19 objects successfully and improve its grasp success rate from 58\% to 65\%. Based on our observations, the reason of failed grasps was that some of the grasp templates were very similar to each other while they represented different types of grasping. 

\subsubsection{Grasping using the approach of Shafii et al.~\cite{shafii2016learning}}

In these experiments, the robot could grasp 17 out of 40 objects in all trials. Included here are all cups (i.e. objects 15, 22, 28, 37), all baskets (i.e. objects 4, 9, 35 and 36), objects 26, 27 which are similar to object no.~3 and also objects no. 20, 29 which are similar to the taught plate (i.e. object no. 5). In these experiments, eight objects were grasped in some trials but not all. It was mainly due to false positives in affordance prediction. For the first round, the overall success rate was 55\% (88 out 160). In the second round, the robot could improve its grasp success rate to 70\% (112 out 160). In this case, the objects no.~12, 14, 25, 31, and 39 were grasped similar to the object no.~8. 

\subsubsection{Grasping using the proposed approach}

In these experiments, the robot successfully grasped 26 out of 40 objects in all four trials. In particular, the robot could grasp all cylindrical objects successfully (i.e., objects 11, 13, 16, 17, 21, 34 and 38) since it recognized the correct affordance and used the right taught grasp template (i.e., the one taught for affordance no. 2). Moreover, the robot could infer that four objects (20, 33, 29, 10) have the same affordance as the plate (i.e., no.~5) and could pick and place them successfully. Similarly, by inferring that another set of objects (i.e., numbers 15, 28, 30, 32 and 37) have the same affordance as the cup (i.e., no.~1), the robot could grasp them successfully in all trials. The robot could also grasp object 8 (pentomino) and objects 22, 26, 27 by using the correct grasp templates taught for the objects no.~3 (spoon) and no.~4 (colander) respectively. The success rate of grasping for all trials was about 65\%. In this round, the robot always failed to grasp 14 objects. For six of them (8, 12, 14, 25, 31 and 39), the affordance was recognized as \emph{Unknown}, and for the remaining eight objects, the grasp points were not detected correctly. 

In the second round, the robot could successfully grasp 38 out of 40 objects in all trials. There were only two objects (i.e., numbers 9 and 36) that the robot failed to grasp. The reason was that the affordance of these objects was not correctly recognized. Since both objects contain lots of holes, the \emph{Object Detection} module could not cluster them properly. As a summary, the robot could improve its success rate from 65\% to 96\%. A video of this experiment can be found online at: {\href{https://youtu.be/MrqmnBbXc70}{\cblue{{https://youtu.be/MrqmnBbXc70}}}

\subsection{System Demonstration}
We also performed two demonstrations to show all the described functionalities of the proposed framework.

\subsubsection{Scene dataset}
We used the Washington RGB-D scene dataset~\cite{henry2013patch} for the first demonstration. This dataset is suitable for this evaluation since it consists of 14 crowded scenes containing several instances of five object categories. In this demonstration, the system initially had prior knowledge about the \emph{Cap}, \emph{Bowl} and \emph{SodaCan} categories, learned from batch data (i.e., a set of observations with ground truth labels), and there is no information about other categories (i.e. \emph{Mug} and \emph{CerealBox}). As depicted in Fig.~\ref{fig:whashington_dataset}~(\emph{left}), the system was able to detect and recognize instances of learned categories and learn new object categories in an online manner. This demonstration is available at: \href{https://youtu.be/yM6nGk_qGHk}{\cblue{{https://youtu.be/yM6nGk\_qGHk}}}

\begin{figure}
\centering
\begin{tabular}[width=\textwidth]{cc}
 \includegraphics[width=0.44\linewidth, trim= 0.5cm 5.4cm 0.5cm 1.3cm,clip=true]{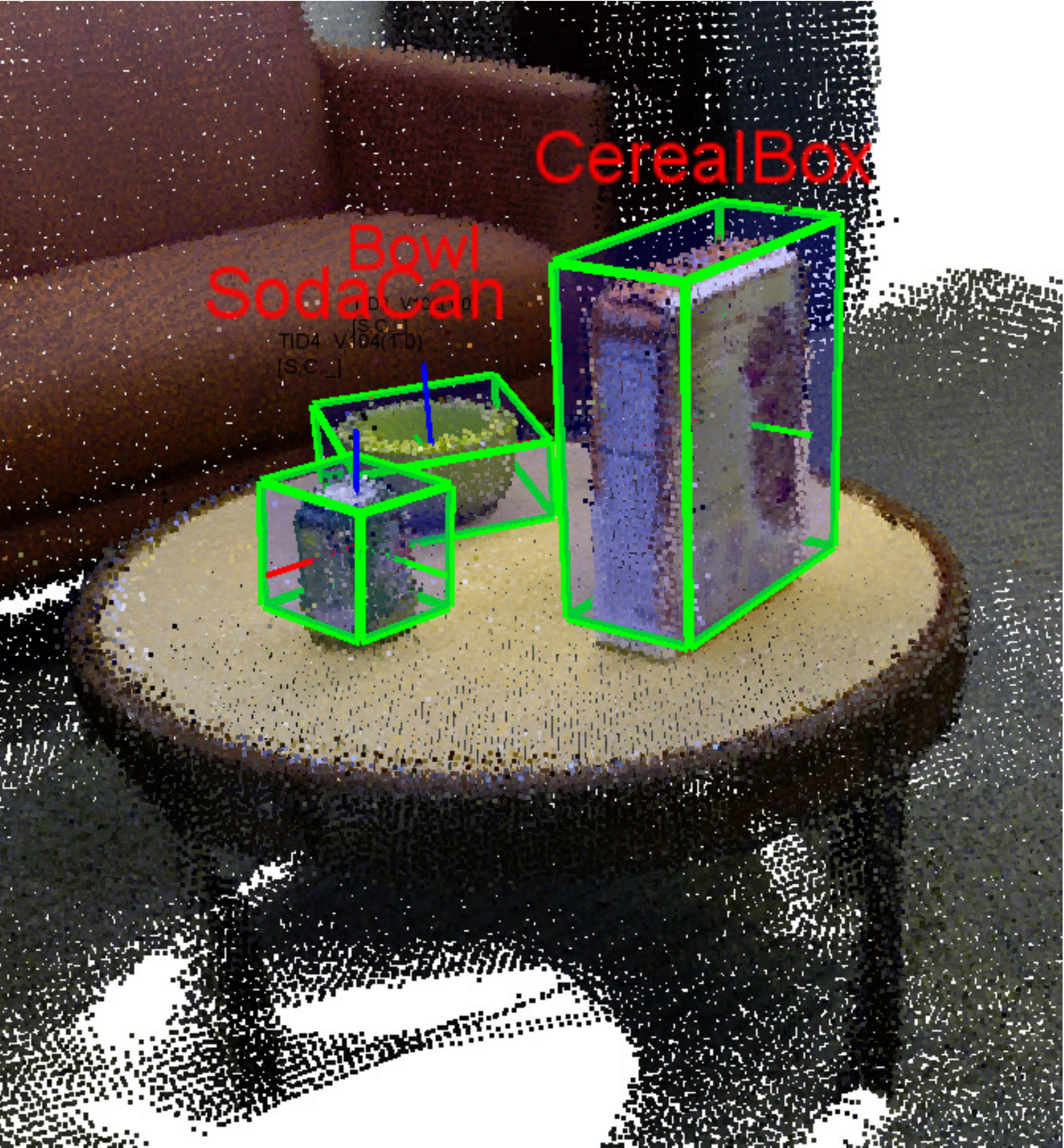} &
  \includegraphics[width=0.48\linewidth, trim= 0cm 0.5cm 0cm 0cm,clip=true]{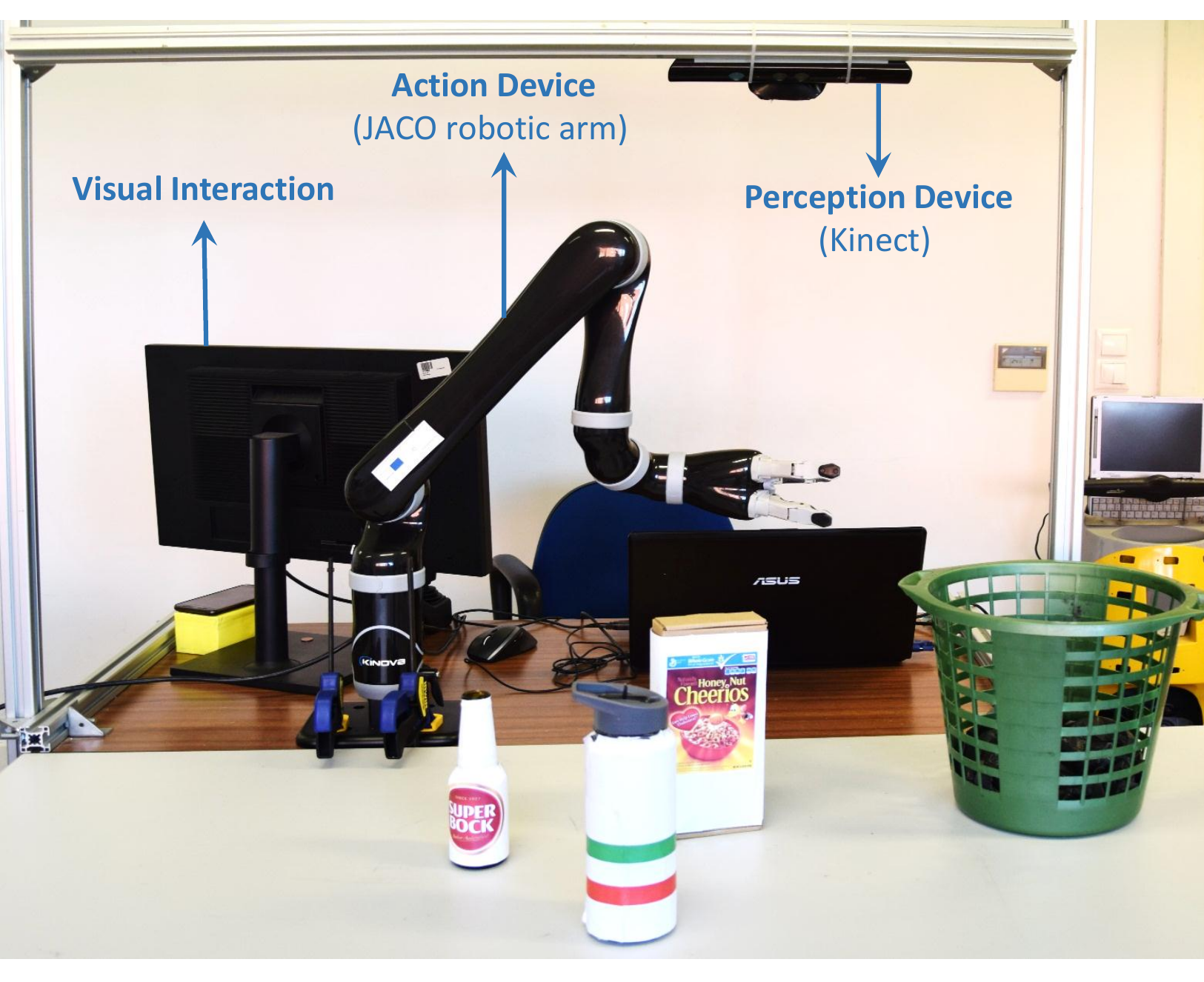} 
\end{tabular}\\
\vspace{0mm}
\caption{ System demonstrations: (\emph{left}) using Washington RGB-D scene dataset~\cite{henry2013patch}; (\emph{right}) real-world robotic application.}
\vspace{-2mm}
\label{fig:whashington_dataset}       
\end{figure}

\subsubsection{Robotic application}
In this demonstration, a user interacts with the system by teaching several objects to the robot and instructing the robot to perform a ``\emph{clear\_table}'' task (Fig.~\ref{fig:whashington_dataset} \emph{right}).  
The system only knew the \emph{TrashBasket} category.  The robot must be able to detect, learn and recognize different objects and transport all objects into the \emph{TrashBasket} to handle this task. While there are objects on the table, the robot retrieves the world model information from the \emph{Working Memory}, including category and position of all active objects. The robot then grasps the nearest object and clears it from the table. A video of this demonstration is online at: \href{https://youtu.be/jYbjGKG4c-U}{\cblue{{https://youtu.be/jYbjGKG4c-U}}}

\section{Conclusions}
In this paper, we have presented a robotic framework includes perception and manipulation capabilities that allow robots to incrementally learn object categories and their affordances from the set of accumulated experiences and reason about how to perform grasping tasks in different situations. To validate our approach, we conducted an extensive set of experiments. Results show that the overall performance of our object and affordance recognition are clearly better than the best results obtained with the state-of-the-art approaches. In the continuation of this work, we will investigate the possibility of using deep transfer learning methods for 3D object recognition in open-ended domains.
Some results obtained with a deep transfer learning approach have already been published~\cite{kasaei2019orthographicnet}.

\section*{ACKNOWLEDGMENT}
{This work was funded by National Funds through the FCT - Foundation for Science and Technology, in the context of the project UID/CEC/00127/2013 and FCT scholarship SFRH/BD/94183/2013.
}


\bibliographystyle{IEEEtran}
\bibliography{library}

\end{document}